\documentclass[%
 aip,
 amsmath,amssymb,
 reprint,%
]{revtex4-1}

\usepackage{graphicx}
\usepackage{dcolumn}
\usepackage{bm}

\usepackage[utf8]{inputenc}
\usepackage[T1]{fontenc}
\usepackage{mathptmx}
\usepackage{etoolbox}
\usepackage{tikz}
\makeatletter
\def\@email#1#2{%
 \endgroup
 \patchcmd{\titleblock@produce}
  {\frontmatter@RRAPformat}
  {\frontmatter@RRAPformat{\produce@RRAP{*#1\href{mailto:#2}{#2}}}\frontmatter@RRAPformat}
  {}{}
}%
\makeatother
\begin{document}

\preprint{AIP/123-QED}

\title[Rethinking Physical Complexity in the Data-Driven Era]{Compression, Regularity, Randomness and Emergent Structure:\\Rethinking Physical Complexity in the Data-Driven Era}
\author{Nima Dehghani}%
    \email{nima.dehghani@mit.edu}
    \homepage{http://compneuro.mit.edu/}
    \affiliation{MIT}%
    \altaffiliation[]{Previously at Department of Physics; Currently at McGovern Institute for Brain Research.}

\date{\today}

\begin{abstract}
Complexity science offers a wide range of measures for quantifying unpredictability, structure, and information. Yet a systematic conceptual organization of these measures is still missing.
We present a unified framework that locates statistical, algorithmic, and dynamical measures along three orthogonal axes—regularity, randomness, and complexity—and situates them in a common conceptual space. We map statistical, algorithmic, and dynamical measures into this conceptual space, discussing their computational accessibility and approximability. This taxonomy reveals the deep challenges posed by uncomputability, and highlights the emergence of modern data-driven methods — including autoencoders, latent dynamical models, symbolic regression, and physics-informed neural networks — as pragmatic approximations to classical complexity ideals. Latent spaces emerge as operational arenas where regularity extraction, noise management, and structured compression converge, bridging theoretical foundations with practical modeling in high-dimensional systems. We close by outlining implications for physics‑informed AI and AI‑guided discovery in complex physical systems, arguing that classical questions of complexity remain central to next‑generation scientific modeling.
\end{abstract}

\maketitle


\section{\label{sec:level1}Introduction}
The quantitative characterization of complexity, regularity, and randomness in empirical data remains a central challenge across the sciences~\cite{Lloyd2001,Prokopenko2009}.

Over the past several decades, a wide range of measures has been developed—spanning statistical entropies, dynamical information flows, algorithmic complexities, and multiscale methods—each probing different facets of structure and unpredictability~\cite{Shannon1948,Kolmogorov1968,Schreiber2000,Costa2005}. Yet, despite this proliferation, a unified framework systematically comparing these measures in terms of the \textit{fundamental properties of data} they capture remains lacking~\cite{Ladyman2013,Shalizi2006}.

Most reviews organize complexity measures taxonomically, either by mathematical category (e.g., Shannon entropy, Kolmogorov complexity, transfer entropy) or by application domain (e.g., neuroscience, finance)~\cite{Lloyd2001,Mitchell2009,Cofre2025}. What is missing is a principled, conceptual framework clarifying precisely how these measures relate to three fundamental properties:

\begin{itemize}
    \item the \textbf{\textit{regularity}} of a system (predictable, structured patterns),
    \item its \textbf{\textit{randomness}} (unpredictable, noisy fluctuations), and
    \item its \textbf{\textit{complexity}} (the interplay between structure and disorder that characterizes non-trivial systems).
\end{itemize}

This paper addresses that gap. We propose a unified perspective that positions existing measures along these three fundamental axes, systematically comparing their operational interpretations, mathematical foundations, and practical implications. Through this framework, we not only organize existing literature but also identify areas requiring new measures or pragmatic approximations, particularly in real-world contexts where idealized theoretical quantities—such as Kolmogorov complexity—are practically inaccessible.

Our approach is motivated by the increasing recognition that regularity and randomness are not opposing absolutes, but deeply intertwined characteristics of complex systems~\cite{Crutchfield2011}. Understanding how various measures dissect these interwoven features is essential for interpreting complex behaviors across diverse fields, from neuroscience and finance to climate science and evolutionary biology~\cite{Goldberger2002,Crutchfield2012}.

In parallel with classical theoretical developments, modern machine learning techniques have significantly advanced data-driven approaches for identifying structure within high-dimensional systems~\cite{LeCun2015, Kingma2014, Brunton2016, Champion2019}. Methods such as autoencoders, variational latent space models, and physics-informed neural networks serve as practical implementations of compression, extracting compact, structured representations from complex, noisy data. These advances underscore the enduring relevance of questions posed herein: ``\textit{how regularity, randomness, and complexity manifest in data?}'', and ``\textit{how they can be comprehensively and systematically characterized?}''.

Thus, the classical measures reviewed in this paper not only clarify conceptual foundations but also provide essential theoretical lenses through which contemporary developments in (neural) data analysis~\cite{Cunningham2014,Dehghani2016,Dehghani2024}, symbolic regression~\cite{Schmidt2009,Cranmer2020,Udrescu2020}, and generative modeling~\cite{Goodfellow2014, Kingma2014, Shol2015, Cao2024} can be better understood.
The paper is structured as follows:

\begin{itemize}
    \item Section II reviews philosophical and mathematical foundations underlying regularity, randomness, and complexity.
    \item Section III introduces our taxonomy of measures, classifying them according to their sensitivity to these fundamental properties.
    \item Section IV presents comparative analyses and visual mappings.
    \item Section V identifies gaps and open problems, and Section VI concludes with directions for future research.
\end{itemize}

\section{Foundations of Regularity, Randomness, and Complexity}
\subsection{Randomness}
Randomness corresponds to unpredictable, noisy fluctuations or the absence of structured order within a system~\cite{Shannon1948,Kolmogorov1968}. Formally, randomness implies a lack of identifiable patterns that would allow prediction or effective compression~\cite{Kolmogorov1968,Martinlof1966,Li2019}.

At its most rigorous, randomness is formalized through \textbf{algorithmic information theory}. A sequence (or string) is considered algorithmically random if its shortest possible description is essentially as long as the sequence itself—this is defined as Kolmogorov–Chaitin complexity~\cite{Kolmogorov1968,Chaitin1969}.

In probabilistic contexts, randomness is captured by \textbf{Shannon entropy}, defined as the expected amount of information (or ``surprise'') associated with observing a new symbol drawn from a given probability distribution~\cite{Shannon1948,Cover2005}.

However, randomness is nuanced. For example, a sequence can exhibit probabilistic randomness—characterized by high Shannon entropy—while still possessing hidden, algorithmically compressible structure, as observed in deterministic chaotic sequences~\cite{Crutchfield2003,Grassberger2012}. Hence, a critical conceptual distinction emerges between \textit{apparent randomness}, characterized by high entropy, and \textit{true randomness}, characterized by high Kolmogorov complexity~\cite{Bialek2001,Li2019}.

\subsection{Regularity}
Regularity refers to predictable, structured patterns observed in the behavior or configuration of a system~\cite{Grassberger1986,Crutchfield2003}. Specifically, regularity indicates the presence of recurring patterns, redundancies, or low-dimensional structures that enable effective prediction or data compression~\cite{Solomonoff1964,Prokopenko2009,Feldman1998}.

Formally, a sequence exhibits regularity if it can be generated by a simple model—characterized by a short description length—and has low conditional entropy given its past~\cite{Crutchfield1989,Rissanen1978}. Regularity thus underlies the capability to construct succinct, predictive models: datasets with significant regularity can typically be summarized by compact rules or governing equations.


Measures of regularity typically quantify predictability, recurrence of patterns, or compressibility~\cite{Bandt2002,Costa2005,Marwan2007}. For example, the entropy rate quantifies residual uncertainty after considering historical information of the observed process; a low entropy rate corresponds directly to high regularity~\cite{Crutchfield2003}.

\subsection{Complexity}
Complexity occupies an intermediate position between the extremes of perfect order and complete randomness~\cite{Gellman1996,Crutchfield1989,Huberman1986}. While a perfectly ordered system (such as a pure sine wave) is predictable but trivial, a completely random system (such as white noise) is unpredictable yet equally uninteresting from a structural standpoint~\cite{Feldman1998}. Complexity emerges in systems characterized by rich structures that are neither trivially predictable nor entirely random~\cite{Gellman1996,Crutchfield1989}.

Several rigorous formalizations of complexity aim to capture this intuitive notion of the ``edge of chaos'':
\begin{itemize}
    \item \textbf{Effective complexity} isolates the description length associated with structural regularities, excluding random fluctuations~\cite{Gellman2004,McAllister2003}.
    \item \textbf{Logical depth} measures complexity as the computational effort required to derive or unfold a compressed representation of a system~\cite{Bennett1988}.
    \item \textbf{Statistical complexity} quantifies the minimal memory required for optimal future prediction of a given process~\cite{Crutchfield1989,Crutchfield1999}.
\end{itemize}

Critically, complexity does not arise merely from high entropy or extensive structure independently; rather, it manifests through a delicate balance between these two properties. Therefore, any comprehensive comparative analysis of complexity measures must explicitly evaluate how effectively each measure captures this essential balance.

\paragraph*{Necessity of Coarse‑Graining.}
Following Lloyd and Pagels~\cite{Lloyd1988}, any complexity measure aiming to differentiate ``regularity'' from ``randomness'' must begin with a clearly defined coarse‑graining map:
\[
  \pi:\Omega_{\mathrm{micro}}\to\Omega_{\mathrm{macro}},
\]
which partitions the complete micro-state space \(\Omega_{\mathrm{micro}}\) into distinct macro‑states \(\Omega_{\mathrm{macro}}\).  Only after establishing such a coarse-graining can one rigorously define the set of regularities:
\[
  R(x)=\pi^{-1}\bigl(\pi(x)\bigr)
  \quad\text{and}\quad
  K\bigl(R(x)\bigr),
\]
ensuring that the identified regularities \(R(x)\) correspond explicitly to macro-level features of interest.

Thus, a foundational step in defining complexity measures is specifying the coarse-graining map \(\pi\), which clearly delineates structural components from random fluctuations.

\subsection{Regularity and Randomness Are Not Opposites}
It may be tempting to view regularity and randomness as opposing extremes on a linear spectrum—assuming that systems must be either highly regular (predictable) or highly random (unpredictable). However, this perspective is misleading~\cite{Crutchfield2011,Crutchfield2003,Prokopenko2009}.

In reality, \textbf{regularity and randomness are independent properties} that often coexist in real-world systems~\cite{Prokopenko2009,Grassberger1986}. For instance, a sequence may contain significant regions of predictable structure embedded within larger random fluctuations, or it might exhibit layers of structure observable only at certain scales or under specific transformations~\cite{Costa2005,Marwan2007}.

This nuanced view has been increasingly recognized within complexity science communities; for example, a Santa Fe Institute (SFI) workshop on ``Randomness, Structure, and Causality: Measures of Complexity from Theory to Applications'' emphasized precisely this attribute: \emph{randomness} and \emph{structure} are fundamentally different concepts and must be treated separately in rigorous analyses~\cite{Crutchfield2011}. While entropy-based metrics primarily quantify unpredictability, algorithmic and statistical complexity measures aim explicitly to isolate structural components obscured by randomness.

Two key reasons underlie this non‑duality:
\begin{itemize}
  \item \textbf{Local Randomness vs.\ Global Structure:}  
     A system can be locally random, meaning individual measurements appear noisy, while simultaneously exhibiting global structural regularities that emerge at macroscopic scales~\cite{Kantz2003}. Chaotic systems provide canonical examples: their trajectories are highly sensitive to initial conditions—demonstrating high local unpredictability—but their underlying attractors possess well-defined, structured low-dimensional geometry.
  \item \textbf{Hidden Regularities:}  
    Not all structural regularities are immediately apparent. Compression algorithms can uncover hidden redundancies that are not obvious upon simple inspection. A dataset may display high entropy at the level of individual symbols yet remain highly compressible once deeper structural rules (e.g., grammatical patterns in natural language) are identified.
\end{itemize}
Thus, \textbf{regularity and randomness are not strict inverses}, but rather \textbf{interwoven attributes}  jointly shaping the informational and structural character of complex datasets.

Critically, interpreting entropy measures requires caution: low entropy indicates predictability but does not guarantee a rich structure, whereas high entropy suggests unpredictability without necessarily excluding meaningful underlying regularities.

\subsection{Compression as a Unifying Lens}
Compression offers a powerful operational framework that naturally unifies the concepts of regularity, randomness, and complexity:

\begin{itemize}
    \item \textbf{Regularity} manifests through \textbf{compressibility}: predictable or repetitive patterns enable concise descriptions that are shorter than the raw data itself~\cite{Cilibrasi2005}.
    \item \textbf{Randomness} resists compression: a truly random sequence possesses no shorter description than the sequence itself~\cite{Kolmogorov1968,Vitanyi2000}.
    \item \textbf{Complexity} emerges in systems that are \textbf{compressible}, indicating structure, yet require \textbf{non-trivial effort} to achieve a succinct description—such systems are neither trivially simple nor fully random~\cite{Bennett1988,Feldman1998}.
\end{itemize}

From the perspective of \textbf{algorithmic information theory}, these relationships can be formalized clearly:

\begin{itemize}
    \item The \textbf{Kolmogorov complexity} \( K(x) \) of a string \( x \) is the length of the shortest program capable of generating \( x \)~\cite{Kolmogorov1968,Vitanyi2000}.
    \item \textbf{Highly regular data} exhibit low \( K(x) \) relative to their original length.
    \item \textbf{Truly random data} have \( K(x) \) values approaching their raw length.
    \item \textbf{Complex data} occupy an intermediate regime, where the shortest program is neither trivial nor as extensive as the original data itself.
\end{itemize}

Thus, a compression-based viewpoint facilitates a nuanced classification of systems:
\begin{itemize}
    \item \textbf{Regular systems:} Highly compressible, governed by simple rules.
    \item \textbf{Random systems:} Minimally compressible, lacking identifiable underlying patterns.
    \item \textbf{Complex systems:} Moderately compressible, characterized by rich, non-trivial structures requiring sophisticated models.
\end{itemize}

This viewpoint also informs practical approaches to complexity measurement. Real-world applications frequently rely on compression algorithms (such as Lempel–Ziv complexity~\cite{Lempel1976,Ziv1978} or gzip compression ratios) as pragmatic proxies for Kolmogorov complexity, albeit recognizing their limitations.

In what follows, we adopt compression—and its intrinsic relationship to predictability and randomness—as a fundamental organizing principle for systematically comparing existing entropy and complexity measures.

This compression-centric perspective naturally extends to contemporary machine learning methodologies. Neural network architectures, notably autoencoders and variational latent models, perform implicit compression by mapping high-dimensional input data into lower-dimensional latent spaces~\cite{Bengio2013,Kingma2014}. These learned latent representations capture the essential regularities (structure), discard irrelevant variations (randomness), and facilitate generative modeling and predictive analyses.

Moreover, emerging paradigms such as physics-informed neural networks (PINNs) and symbolic regression explicitly aim to identify compact, interpretable governing rules from complex data. These approaches embody the same core principle of seeking minimal, structured representations underpinning the concept of effective complexity.

Thus, compression serves not only as a theoretical bridge uniting regularity, randomness, and complexity but also as a practical foundation driving the ongoing evolution of data-driven science.

\section{Taxonomy of Complexity Measures}
\subsection{Statistical Measures}

\subsubsection{Overview}
Statistical entropy measures quantify uncertainty, unpredictability, or diversity within the distribution of observed outcomes~\cite{Shannon1948}. They primarily address the question of \textbf{how unpredictable or disordered} a system's outputs appear, typically without explicitly invoking an underlying generative mechanism or structural model~\cite{Prokopenko2009,Paninski2003}.

In general, these measures (Table~\ref{tab:summary_statistics}):

\begin{itemize}
    \item \textbf{Strongly capture randomness}, where higher values indicate greater unpredictability.
    \item \textbf{Weakly capture regularity}, primarily reflected indirectly through lower entropy values.
    \item \textbf{Capture complexity}, only in limited ways (as an incidental balance between low and high entropy--reflecting a balance between order and disorder).
\end{itemize}

Statistical entropy measures operate directly on empirical distributions or observed sequences, making them computationally accessible and widely applicable—particularly useful for analyzing short, noisy datasets where comprehensive model-based reconstruction may not be feasible.

Modern extensions using neural-network-based approaches (such as neural entropy estimators or variational approximations) have expanded classical entropy estimation into higher-dimensional spaces, indicating a growing convergence between traditional statistical measures and contemporary machine learning methodologies.

\begin{table*}
\footnotesize
\caption{\label{tab:summary_statistics} \textbf{Statistical Measures}: Comparison of statistical entropy measures based on their capability to capture randomness, regularity, and complexity.}
\begin{ruledtabular}
\begin{tabular}{lcccc}
\textbf{Measure} & \textbf{Captures Randomness} & \textbf{Captures Regularity} & \textbf{Captures Complexity} & \textbf{Key Mechanism} \\
\hline
Shannon Entropy & Yes & No & No & Average information per symbol \\
Renyi Entropy & Yes (tunable) & No & No & Generalized order-$\alpha$ entropy \\
Tsallis Entropy & Yes (tunable) & No & No & Non-extensive entropy \\
Approximate Entropy (ApEn) & Yes (approximate) & Yes (weak) & Approximate & Pattern recurrence likelihood \\
Sample Entropy (SampEn) & Yes (approximate) & Yes (weak) & Approximate & Improved pattern recurrence measure \\
Permutation Entropy (PE) & Yes (via ordinal patterns) & Some (order structure) & Approximate & Ordinal (rank-order) pattern entropy \\
\hline
\multicolumn{5}{c}{\textbf{Strengths}} \\
\hline
Shannon Entropy & \multicolumn{4}{l}{Canonical, conceptually clear} \\
Renyi Entropy & \multicolumn{4}{l}{Tunable sensitivity to event frequencies} \\
Tsallis Entropy & \multicolumn{4}{l}{Captures non-extensive, heavy-tailed behavior} \\
Approximate Entropy (ApEn) & \multicolumn{4}{l}{Practical, robust to noise} \\
Sample Entropy (SampEn) & \multicolumn{4}{l}{Reduced bias compared to ApEn} \\
Permutation Entropy (PE) & \multicolumn{4}{l}{Robust to noise and monotonic distortions} \\
\hline
\multicolumn{5}{c}{\textbf{Weaknesses}} \\
\hline
Shannon Entropy & \multicolumn{4}{l}{Ignores structural patterns beyond frequency} \\
Renyi Entropy & \multicolumn{4}{l}{Sensitive to choice of parameter $\alpha$} \\
Tsallis Entropy & \multicolumn{4}{l}{Interpretation depends strongly on parameter $q$} \\
Approximate Entropy (ApEn) & \multicolumn{4}{l}{Sensitive to parameter choices} \\
Sample Entropy (SampEn) & \multicolumn{4}{l}{Performance declines with short datasets} \\
Permutation Entropy (PE) & \multicolumn{4}{l}{Ignores magnitude information} \\
\end{tabular}
\end{ruledtabular}
\end{table*}

\subsubsection{Detailed Measure Summaries}

\paragraph{\textbf{Shannon Entropy}}
Shannon entropy,  as
\[
H(X) = -\sum_{i} p(x_i) \log p(x_i)
\]
quantifies the expected information content (or ``surprise'') of an outcome drawn from a distribution~\cite{Shannon1948,Cover2005}. Shannon entropy is the foundational measure of statistical randomness. High value implies unpredictability, whereas low entropy indicates predictability. However, Shannon entropy is \textit{insensitive to structural regularities} beyond simple symbol frequencies and thus does not strongly capture complexity.

\paragraph{\textbf{Renyi Entropy}}
Rényi entropy generalizes Shannon entropy by introducing an order parameter $\alpha$, enabling adjustment of sensitivity toward rare or common events~\cite{Renyi1961}:

\[
H_{\alpha}(X) = \frac{1}{1 - \alpha} \log \sum_{i} p(x_i)^{\alpha}
\]
The parameter $\alpha$ emphasizes different aspects of the distribution (with $\alpha > 1$ emphasizing common events and $\alpha < 1$ highlighting rare events). This flexibility makes Rényi entropy useful for analyzing skewed or heavy-tailed distributions, although interpretation depends sensitively on the chosen value of $\alpha$.

\paragraph{\textbf{Tsallis Entropy}}
Tsallis entropy modifies the Shannon entropy formula to incorporate non-extensive behavior:

\[
S_q(X) = \frac{1 - \sum_{i} p(x_i)^q}{q - 1}
\]

It is particularly suited for systems exhibiting non-equilibrium dynamics, anomalous diffusion, or long-range correlations~\cite{Tsallis1988,Tsallis2003}. The tunable parameter $q$ adjusts its sensitivity to the probability distribution, but its interpretation requires careful consideration of the underlying physics or dynamics.

\paragraph{\textbf{Approximate Entropy (ApEn)}}
Approximate entropy quantifies regularity and unpredictability in time series data by assessing the likelihood that similar patterns persist over consecutive time steps~\cite{Pincus1991,Pincus1996}. ApEn depends on a tolerance parameter \( r \) and embedding dimension \( m \), and counts matching templates of length \( m \) and \( m+1 \), helping to measure how much unpredictability exists in a system~\cite{Pincus1991,Pincus1996}. Formally,
\[
ApEn(m, r, N) = \Phi(m, r) - \Phi(m+1, r)
\]

where

\[
\Phi(m, r) = \frac{1}{N-m} \sum_{i=1}^{N-m} \log C_i^m(r)
\]
and \( C_i^m(r) \) is the fraction of template vectors of length \( m \) that remain close within the tolerance \( r \). ApEn is practical for short, noisy datasets but is sensitive to parameter choices and can exhibit biases due to self-matching.

\paragraph{\textbf{Sample Entropy (SampEn)}}
Sample entropy improves upon ApEn by excluding self-matches, thereby reducing bias and enhancing reliability for short time series~\cite{Richman2000,Lake2002}, making it a popular choice in physiology (e.g., heart rate variability analysis)~\cite{Richman2000,Lake2002}. It is defined as:

\[
SampEn(m, r, N) = -\log \frac{A}{B}
\]

where

\( A \) counts pairs of sequences of length \( m+1 \) matching within tolerance \( r \), and \( B \) is the number of pairs of sequences of length \( m \) that match within the same tolerance \( r \).

By avoiding self-matching, SampEn provides a more unbiased measure of complexity compared to ApEn. However, like ApEn, SampEn depends on parameter choices (embedding dimension \( m \), tolerance \( r \)) and can be unstable with very short time series.

Compared to Shannon or Rényi entropy, SampEn specifically captures dynamical regularity rather than overall distribution uncertainty. Its focus on temporal structure makes it valuable for analyzing non-stationary signals, such as neural activity~\cite{Jia2017,Delgado2019,Cofre2025} or cardiac rhythms~\cite{Goldberger2002}.

To address limitations in short-term sensitivity, a multiscale extension called, called Multiscale Sample Entropy (MSE), applies coarse-graining to time series before computing entropy. MSE improves sensitivity to long-range dependencies and structured variability, making it particularly useful in physiological and complex systems analysis~\cite{Costa2002,Costa2005,Costa2015}.

\paragraph{\textbf{Permutation Entropy (PE)}}
Permutation entropy evaluates the unpredictability of a time series by examining the frequency of ordinal patterns, defined by rank-order sequences of length 
\(m\), computing Shannon entropy over their occurrence~\cite{Bandt2002,Zanin2012}. It is computed as:

\[
PE(m) = -\sum_{\pi} p(\pi) \log p(\pi)
\]

where:

$\pi$ represents distinct ordinal patterns of length $m$, $p(\pi)$ is the relative frequency of occurrence of each pattern in the time series. PE is robust against monotonic nonlinear transformations and amplitude distortions, making it effective for detecting dynamical transitions. However, its insensitivity to magnitude information can lead it to overlook some structural features present in the original signal.

Unlike SampEn and ApEn, which analyze temporal dependencies, Permutation Entropy focuses on the rank order of values, making it less sensitive to amplitude variations. While this makes PE highly effective for detecting state transitions in dynamical systems, it may overlook fine-scale structural complexity that SampEn and ApEn capture~\cite{Zanin2012,Riedl2013}.

Just as Sample Entropy has multiscale extensions that apply coarse-graining to account for long-range dependencies and structured variability, Permutation Entropy also has multiscale variants. These adaptations enhance its ability to detect complex dynamical patterns while preserving robustness against amplitude distortions, making it particularly useful for analyzing hierarchical and multi-scale systems~\cite{Aziz2005,Azami2016,Davalos2019,Morel2021}.

\subsection{Algorithmic Measures}
\subsubsection{Overview}
Algorithmic complexity measures focus not on empirical frequencies or probability distributions but rather on the \textbf{minimal description length} required to generate an object or dataset~\cite{Grunwald2007,Rissanen1978}. Rooted in \textbf{algorithmic information theory}, these measures seek to capture the intrinsic, deep structure within data, explicitly addressing (Table~\ref{tab:algorithmic_measures}):

\begin{itemize}
    \item The degree to which data can be \textbf{compressed}.
    \item The minimal \textbf{rules} or \textbf{program} necessary for its reproduction.
    \item The associated \textbf{computational effort}.
\end{itemize}

In general, algorithmic measures:

\begin{itemize}
    \item \textbf{Strongly capture regularity}, since structured patterns enable compression.
    \item \textbf{Strongly capture randomness}, since randomness inherently resists compression.
    \item \textbf{Directly characterize complexity} as the interplay between structured and unstructured components.
\end{itemize}

While theoretically robust, pure algorithmic measures (such as Kolmogorov complexity) are typically \textbf{uncomputable} or computationally prohibitive. Therefore, practical proxies (e.g., compression algorithms, model-selection criteria, and neural network approximations) have been developed for real-world analyses.


These algorithmic measure ideas align strongly with modern machine learning approaches, where neural architectures like autoencoders, generative models, and symbolic regression techniques implicitly perform learned data compression, approximating minimal representations of complex structures.

\begin{table*}
\footnotesize
\caption{\label{tab:algorithmic_measures}\textbf{Comparison of Complexity Measures}: Algorithmic complexity measures and their characteristics; \textbf{shaded cells highlight uncomputable ideals}.}
\begin{ruledtabular}
\begin{tabular}{lccc}
\textbf{Measure} & \textbf{Captures Randomness} & \textbf{Captures Regularity} & \textbf{Captures Complexity} \\
\hline
Kolmogorov Complexity & Yes & Yes & Yes (indirect) \\
Effective Complexity & No & Yes & Yes (direct) \\
Logical Depth & No & Yes & Yes \\
Sophistication & No & Yes & Yes \\
\end{tabular}
\end{ruledtabular}

\vspace{5mm} 

\begin{ruledtabular}
\begin{tabular}{p{4cm} p{11cm}} 
\textbf{Measure} & \textbf{Key Mechanism} \\
\hline
Kolmogorov Complexity & Shortest program length \\
Effective Complexity & Description of structured regularities \\
Logical Depth & Computation time of minimal program \\
Sophistication & Size of minimal sufficient statistic \\
\end{tabular}
\end{ruledtabular}

\vspace{5mm}

\begin{ruledtabular}
\begin{tabular}{p{4cm} p{11cm}} 
\textbf{Measure} & \textbf{Strengths} \\
\hline
Kolmogorov Complexity & Theoretical gold standard \\
Effective Complexity & Separates structure from noise \\
Logical Depth & Captures computational richness \\
Sophistication & Quantifies richness of structure \\
\end{tabular}
\end{ruledtabular}

\vspace{5mm}

\begin{ruledtabular}
\begin{tabular}{p{4cm} p{11cm}} 
\textbf{Measure} & \textbf{Weaknesses} \\
\hline
Kolmogorov Complexity & Uncomputable; practical approximations required \\
Effective Complexity & Boundary between structure/noise hard to precisely define \\
Logical Depth & Difficult to approximate \\
Sophistication & Subtle theoretical interpretation, unstable in practice \\
\end{tabular}
\end{ruledtabular}
\end{table*}

\subsubsection{Detailed Measure Summaries}
\paragraph{\textbf{Kolmogorov Complexity}}

Kolmogorov complexity \( K(x) \) of a string \( x \) is the length of the shortest program that outputs \( x \) on a universal Turing machine~\cite{Kolmogorov1968,Li2019}. It provides an absolute, model-free measure of an object's intrinsic informational content:

\begin{itemize}
    \item Highly regular data exhibit low \( K(x) \).
    \item Truly random data approach \( K(x) \) close to the length of \( x \) itself.
\end{itemize}

However, Kolmogorov complexity is inherently \textbf{uncomputable}, necessitating the use of compression algorithms (e.g., Lempel-Ziv complexity, deep generative models) as practical approximations.

\paragraph{\textbf{Logical Depth}}

Logical Depth, proposed by Bennett~\cite{Bennett1988,Bennett1995}, quantifies complexity in terms of the computational effort required to generate an object or dataset from its shortest description. It formalizes the intuition that genuinely complex structures are computationally ``deep''—they require nontrivial computational resources to unfold, unlike trivial or purely random structures.

Formally, Bennett’s logical depth of a string \(x\) at significance level \(\nu\) is defined as:
\[
  \mathrm{Depth}_\nu(x)
  = \min\Bigl\{\,t : \exists\,p,\; |p|\le K(x)+\nu,\;
    \mathcal{U}(p)\xrightarrow{t} x\Bigr\},
\]
where:
\begin{description}
    \item \(\mathcal{U}\) denotes a universal Turing machine,
    \item \(k(x)\) is the Kolmogorov complexity of \(x\),
    \item \(\nu\) sets a significance threshold allowing programs slightly longer than the minimal one,
    \item \(t\) is the minimal computation time (number of steps) needed to produce \(x\) from a valid program \(p\) whose length \(|P|\) is within \(\nu\) bits of the minimal complexity \(K(x)\).
\end{description}

In simplified terms, Logical Depth can be expressed heuristically as:

\[
LD(x) = \text{Time}(P_x)
\]

where:
\begin{description}
    \item \( P_x \) is the shortest program that outputs \( x \) on a universal Turing machine,
    \item \( \text{Time}(P_x) \) is the execution time required to run \( P_x \).
\end{description}

This formalization clarifies that random strings, despite their incompressibility, are computationally shallow: their minimal programs simply encode the exact sequence without meaningful computational unfolding. Conversely, structured data with rich complexity are typically computationally deep, requiring substantial computation from highly compressed descriptions.

Since Logical Depth involves both the minimal program length and computational runtime, exact calculation is generally infeasible for complex datasets. Thus, practical approximations are often employed, using standard compression algorithms as heuristics. One such heuristic measure is the wall-clock runtime of decompressing a compressed representation:
\[
  \widehat{\mathrm{Depth}}(x)
  = \mathrm{time}\bigl(\mathrm{decompress}(\mathrm{compress}(x))\bigr).
\]

\paragraph{\textbf{Effective Complexity}}
Having described Kolmogorov complexity and logical depth as foundational measures of complexity, we now turn to thermodynamic depth, introduced by Lloyd and Pagels, as a complementary perspective and a precursor to Effective Complexity. Both logical depth and thermodynamic depth relate to the history of a system, but they do so in distinct ways—\textit{logical depth is computational}: A system with high logical depth requires significant processing time to generate, meaning it has a rich causal history. While \textit{thermodynamic depth focuses on the physical irreversibility of a system’s formation}: it quantifies the amount of thermodynamic work needed to generate a system from a random initial state. 

\textit{Thermodynamic Depth  (related to Effective Complexity):}
Lloyd and Pagels~\cite{Lloyd1988} define the \emph{thermodynamic depth} of a macro‑state \(\sigma\) as
\[
  D_{\mathrm{thermo}}(\sigma)
  = \sum_{\substack{\text{micro‑trajectories }\tau: \\ \tau(0)\to\sigma}}
    P(\tau)\,\Delta S[\tau],
\]
where 
\begin{description}
    \item  \(\Delta S[\tau]\) denotes the entropy dissipated along trajectory \(\tau\),
    \item \(P(\tau)\) is the probability of trajectory \(\tau\).
\end{description}

This quantity measures the minimal free-energy or entropy cost to assemble a given macroscopic state, complementing logical depth’s informational perspective from a thermodynamic perspective~\cite{Lloyd1988,Crutchfield1999}.

Building upon these ideas, Effective Complexity (EC), introduced by Gell-Mann and Lloyd~\cite{Gellman1996,Gellman2004}, provides another dimension. While thermodynamic depth quantifies the physical effort required to generate a system, effective complexity focuses on the structured information within it. EC refines the Kolmogorov definition of complexity by explicitly isolating the complexity arising solely from the structured regularities in a dataset. Thus, pure randomness has low Effective Complexity despite having high Kolmogorov complexity.

A crucial distinction is that Effective Complexity does not measure disorder or randomness—\textit{it specifically quantifies the complexity associated with meaningful structure within a system}. This makes it uniquely relevant for analyzing real-world phenomena where structured patterns coexist with noise~\cite{Gellman1996,Gellman2004,McAllister2003}. It is formally given by the Kolmogorov complexity of the simplest description of these regularities:


\[
  \mathcal{E}(x)=\min_{S:x\in S}\{\,K(S)\colon K(x\mid S)\le\alpha\},
\]

where:

\begin{description}
  \item[\(K(S)\)] is the Kolmogorov complexity of the structured model \(S\),
  \item[\(K(x\mid S)\)] is the conditional Kolmogorov complexity of the data \(x\) given \(S\),
  \item[\(\alpha\)] is a threshold parameter that sets how well the model must capture the regularities in \(x\).
\end{description}

To reinforce the intuition behind effective complexity, it can be informally expressed as
\[
EC(x) = K(R(x))
\]

\begin{description}
    \item[\( K(R(x)) \)] is the Kolmogorov complexity of the regularities \( R(x) \) in the dataset \( x \),
    \item[\( R(x) \)] represents the structured, non-random components of \( x \).
\end{description}

Because exact Kolmogorov complexity \(K(\cdot)\) is uncomputable, practical approximations rely on standard compression algorithms (e.g., LZMA, BZIP2) to estimate description lengths. A common proxy is:

\[
  \widehat{\mathcal{E}}(x)
  = \min_{S}\bigl\{\mathrm{len}(\mathrm{compress}(S)) 
    \;:\;\mathrm{len}(\mathrm{compress}(S,x))\le \beta\bigr\}.
\]

These compression-based approximations are useful because real-world data often contains structured redundancies that compressors can identify. Although imperfect, such methods provide a tractable way to approximate effective complexity in practical settings.

In essence, Effective Complexity prioritizes \textbf{structured richness}: systems characterized by meaningful, complex patterns (such as neural data, genomes or natural languages) possess high effective complexity, even amidst noisiness. Since irreversible transformations shape the structured components of a system, thermodynamic depth can contribute to the emergence of meaningful regularities. In this sense, systems with high thermodynamic depth may provide a foundation for high effective complexity, as structured patterns often arise through physical processes that retain history.

\paragraph{\textbf{Statistical Complexity}}
Statistical Complexity $C_\mu$, introduced by Crutchfield and Young~\cite{Crutchfield1989,Lizier2014,Crutchfield2017}, quantifies complexity as the minimal information required to optimally predict future observations from past data. It is formally defined as the Shannon entropy of the distribution over the causal states identified by an $\epsilon$-machine:

\[
  C_\mu = H[\mathcal{S}],
  \quad \mathcal{S}=\epsilon(\mathcal{P}(x)),
\]

where \(\mathcal{P}(x)\)  represents all past observations of \(x\) leading up to the present state, and the function \(\epsilon(\mathcal{P}(x))\) groups all such pasts that induce identical predictive distribution.

This entropy quantifies the total predictive information stored in the system’s causal structure, ensuring that future states can be optimally inferred from past data. Thus, the causal states of an $\epsilon$‑machine provide a data-driven coarse-graining of the observed process—effectively generating \emph{``macrostates''} \(\mathcal{S}\) that represent the minimal sufficient statistics for prediction. This closely parallels the idea of Effective Complexity, which isolates regularities \(R(x)\) as the macro-features essential for description. Both Statistical Complexity and Effective Complexity therefore aim to quantify the amount of structured, non-random information needed to optimally forecast or reconstruct system behavior. Equivalently, one may interpret \(C_\mu\) as the coding cost (in bits) of this minimal predictive representation.

Beyond merely defining macrostates, the $\epsilon$‑machine embodies an ``artificial science'' approach~\cite{Crutchfield2012}, in which structural models emerge inductively from data rather than being imposed \emph{a priori}. This approach ensures that identified causal states represent intrinsic structure rather than arbitrary modeling assumptions. While Effective Complexity identifies structured regularities for description, Statistical Complexity determines the minimal information necessary to retain for accurate prediction. In this sense, \textit{Effective Complexity is structure-focused, while Statistical Complexity is prediction-focused}.

\noindent\emph{Practical note:} Empirical reconstruction of $\epsilon$‑machine (e.g., using algorithms such as CSSR~\cite{Shalizi2004}) can sometimes over-split causal states when data are limited, resulting in upward-biased estimates of \(C_\mu\).  To address this issue, regularization methods—such as penalizing the number of states or employing Bayesian state-merging techniques—can encourage parsimonious models and avoid counting spurious structures as meaningful complexity.

\paragraph{\textbf{Sophistication}}
Sophistication~\cite{Koppel1987,Antunes2013} measures complexity as the size of the \textbf{minimal sufficient statistic} required to describe a given object or dataset \(x\). 
Intuitively, sophistication quantifies the minimal complexity of the simplest model that captures meaningful structure in \(x\), while avoiding overfitting to noise. Thus, sophistication balances model complexity (size) against the data-fit criterion, distinguishing structural richness from irrelevant or random details.

Formally, Koppel’s definition of a string \(x\) at significance level \(\beta\) is given by:
\[
  \mathrm{Soph}_\beta(x)
  = \min \Bigl\{\,|S| : S \ni x,\; K(S) + \log|S| \le K(x) + \beta \Bigr\},
\]
where
\begin{description}
  \item \(S\) is a finite set (model) containing the observed object/data \(x\),
  \item \(|S|\) is the cardinality (size) of the model \(S\),
  \item \(K(S)\) is the Kolmogorov complexity of the model \(S\),
  \item \(K(x)\) is the Kolmogorov complexity of object/data \(x\),
  \item \(\beta\) is a slack parameter allowing near‑optimal models, controlling the trade-off between model simplicity and fit accuracy.
\end{description}

Sophistication thus formally captures the idea that meaningful complexity resides in the simplicity of a sufficient explanatory model rather than in randomness itself. Data explained adequately by smaller models exhibit lower sophistication, while those requiring larger, structurally richer models have higher sophistication. A key insight is that smaller models are preferable when they sufficiently explain structure, as excessive complexity often indicates overfitting to noise rather than true sophistication.

Thus, Sophistication measures the complexity of the best explanatory model, while Effective Complexity quantifies the intrinsic complexity of structured patterns within a dataset. One is about model sufficiency, and the other is about structural depth.

In theoretical settings, sophistication provides a rigorous measure, but real-world applications require practical methods for estimating it. However, estimating sophistication precisely can be nuanced. In practice, it is sensitive to technical choices regarding what constitutes a sufficient statistic or an adequate model. Consequently, practical approximations often rely on well-known model-selection criteria such as the Minimum Description Length (MDL) principle or Akaike Information Criterion (AIC), balancing the description length \(K(S)\) against data‑fit \(\log|S|\).   
 
\subsection{Dynamical Information Measures}
\subsubsection{Overview}
Dynamical information measures extend the analysis of complexity to explicitly account for \emph{temporal dependencies}, \emph{causal relationships}, and \emph{information flow} over time. They address not only what information is present but also how it is processed, stored, and transmitted between components of a system. These measures (Table~\ref{tab:information_measures}):

\begin{itemize}
    \item \textbf{Capture randomness} (via unpredictability of future states),
    \item \textbf{Capture regularity} (via persistence of information from the past),
    \item \textbf{Capture complexity} (via structured dependencies and organized flows of information).
\end{itemize}

Echoing the SFI workshop on ``Randomness, Structure, and Causality''~\cite{Crutchfield2011}, dynamical complexity arises from the interplay between driving forces toward regularity (e.g., coherent patterns, strong coupling) and countervailing randomness (e.g., stochastic noise, high-dimensional chaos). Measures such as transfer entropy and active information storage sit at this boundary, quantifying emergent organization that neither extreme alone would produce.

Importantly, these measures connect naturally to modern machine-learning approaches—state-space models, recurrent neural networks (RNNs), autoencoders of neural dynamics, and Koopman-based latent dynamics all aim to \textit{capture structured information flow} in evolving systems. Thus, dynamical information measures bridge classical information theory and contemporary data-driven methods for learning and interpreting system dynamics.

\begin{table*}
\footnotesize
\caption{\label{tab:information_measures}\textbf{Comparison of Information-Theoretic Measures}: Measures probing storage, transfer, and modification of information over time.}
\begin{ruledtabular}
\begin{tabular}{lccc}
\textbf{Measure} & \textbf{Captures Randomness} & \textbf{Captures Regularity} & \textbf{Captures Complexity} \\
\hline
Entropy Rate & Yes & Yes (via memory) & Approximate \\
Transfer Entropy & Yes & Yes (cross-variable) & Some \\
Active Information Storage & Yes & Yes (within-variable) & Some \\
Information Modification & No (direct) & Yes & Yes (high) \\
\end{tabular}
\end{ruledtabular}

\vspace{5mm} 

\begin{ruledtabular}
\begin{tabular}{p{4cm} p{11cm}} 
\textbf{Measure} & \textbf{Key Mechanism} \\
\hline
Entropy Rate & Uncertainty per symbol given past \\
Transfer Entropy & Information flow $X_t \rightarrow Y_{t+1}$ \\
Active Information Storage & Predictive information from past \\
Information Modification & Synergy and nontrivial interactions \\
\end{tabular}
\end{ruledtabular}

\vspace{5mm}

\begin{ruledtabular}
\begin{tabular}{p{4cm} p{11cm}} 
\textbf{Measure} & \textbf{Strengths} \\
\hline
Entropy Rate & Captures predictability structure \\
Transfer Entropy & Detects directed influence \\
Active Information Storage & Quantifies internal memory \\
Information Modification & Captures emergent computation \\
\end{tabular}
\end{ruledtabular}

\vspace{5mm}

\begin{ruledtabular}
\begin{tabular}{p{4cm} p{11cm}} 
\textbf{Measure} & \textbf{Weaknesses} \\
\hline
Entropy Rate & Hard to estimate in high dimensions \\
Transfer Entropy & Requires large data; estimation bias risks \\
Active Information Storage & Sensitive to noise, nonstationarity \\
Information Modification & Definitions and estimators still maturing; hard to define precisely \\
\end{tabular}
\end{ruledtabular}
\end{table*}

\subsubsection{Detailed Measure Summaries}
\paragraph{\textbf{Entropy Rate}}
Entropy rate \( h_{\mu} \) quantifies the average uncertainty per symbol conditioned on the entire past~\cite{Cover2005,Crutchfield2003}:

\[
h_{\mu} = \lim_{n \to \infty} \frac{1}{n} H(X_1, X_2, \dots, X_n)
\]

\begin{itemize}
    \item A low entropy rate \( h_{\mu} \) implies strong regularity and memory.
    \item A high entropy rate  \( h_{\mu} \) implies weak temporal structure or high randomness.
\end{itemize}

Entropy rate is fundamental in ergodic and dynamical systems theory but can be challenging to estimate accurately, especially in high-dimensional or finite-data settings.

\paragraph{\textbf{Transfer Entropy}}
Transfer entropy \(T_{X \to Y}\) measures the directed, time-asymmetric information flow from one process \( X \) to another process \( Y \)~\cite{Schreiber2000,Hlavackova2007,Vicente2011}:

\[
T_{X \to Y} = H(Y_{t+1} \mid Y_t) - H(Y_{t+1} \mid Y_t, X_t)
\]

This quantity captures the reduction in uncertainty about \( Y_{t+1} \) provided by \( X_t \) beyond the information contained in \( Y_t \) alone. Unlike mutual information, transfer entropy is explicitly directional and sensitive to nonlinear dependencies. It has been widely used in neuroscience (especially for inferring functional connectivity from physiological measurements~\cite{Vicente2011,Orlandi2014,Wibral2014,Lobier2014,Novelli2019}), finance (with a focus on quantifying market dependencies and assessing risk contagion\cite{Marschinski2002,Kwon2008,Dimpfl2013}), and climatology (mostly for detecting flow of influence and connections between different locations or climate variables to infer causal-like interactions~\cite{Campuzano2018,Bennett2019,Delgado2020,Tongal2021, Silini2023}). However, reliable estimation of transfer entropy requires large datasets and careful bias correction, particularly in high-dimensional systems.

Entropy rate and transfer entropy are closely related measures of information flow and predictability. \textit{Entropy rate} quantifies the intrinsic uncertainty in a process, while \textit{transfer entropy} evaluates how much uncertainty reduction occurs when incorporating an external source of information. A system with \emph{low entropy rate} exhibits strong regularity and predictable behavior, meaning transfer entropy may reveal structured dependencies between processes. Conversely, in \emph{high-entropy systems}, transfer entropy can help identify dominant information flows amidst apparent randomness.

\paragraph{\textbf{Active Information Storage}}
Active Information Storage (AIS) quantifies how much information about the future state of a variable is contained within its own past~\cite{Lizier2014,Wibral2013}:

\[
AIS = I(X_{\text{past}} ; X_{\text{future}})
\]
where \( I(\cdot ; \cdot) \) denotes mutual information. AIS measures a system’s internal memory—how past states predict future ones. It can be useful for analyzing neural dynamics and adaptive systems but can be sensitive to noise and embedding choices.

Consider a dynamical system where one variable influences another through directional coupling. In such cases, Transfer Entropy (TE) quantifies how much information flows between the variables, whereas \(AIS\) measures how much predictive information is stored within the system itself. If the coupling is strong, TE tends to exceed AIS, demonstrating that external influence enhances predictability beyond intrinsic memory.

\paragraph{\textbf{Information Modification}}
Information Modification captures the emergence of new information through synergistic interactions among multiple sources. It is particularly concerned with \textbf{nonlinear, collective influences} where multiple inputs combine to produce outcomes that cannot be attributed to individual sources alone~\cite{Lizier2010,Williams2010}.

A formal approach to quantifying this phenomenon relies on Partial Information Decomposition (PID), which separates unique, redundant, and synergistic contributions to a target variable. The mutual information \(I(Y; \{X_1, X_2\})\) is decomposed into:

\begin{align}
I(Y; \{X_1, X_2\}) &= I_{\text{unique}}(Y; X_1) + I_{\text{unique}}(Y; X_2) \notag \\
&\quad + I_{\text{redundant}}(Y; X_1, X_2) + I_{\text{synergistic}}(Y; X_1, X_2)
\end{align}

where:
\begin{description}
    \item [\( I_{\text{unique}}(Y; X_i) \)] The information provided uniquely by \( X_i \) about \( Y \).
    \item [\( I_{\text{redundant}}(Y; X_1, X_2) \)] The shared (overlapping) information available from both sources.
    \item [\( I_{\text{synergistic}}(Y; X_1, X_2) \)] The new information that emerges \textbf{only when both sources are considered together}.
\end{description}

Information Modification attempts to quantify true emergent computation—new structure arising solely through interactions. This perspective has been explored in neuroscience, complex systems, and computational biology to understand higher-order dependencies~\cite{Williams2010,Chan2017,Rosas2020,Sherrill2021}. Unlike Transfer Entropy, which measures direct information flow, Information Modification captures cases where information is created through synergistic interactions. TE tracks directional influence, while Information Modification reveals emergent structure that only arises when multiple inputs act jointly.

Despite its conceptual richness, Information Modification remains an evolving field with open questions regarding operational definitions and estimation methods. Practical implementations require careful choices of decomposition frameworks, especially in high-dimensional or noisy systems.

\section{Conceptual Maps and Computability}
\subsection{The Regularity–Randomness–Complexity Landscape}
In earlier sections we established \textbf{regularity}, \textbf{randomness}, and \textbf{complexity} as three distinct yet intertwined properties of data. This reflects the emerging notion that randomness and regularity are not the exact opposites ~\cite{Crutchfield2012}. Together, \emph{regularity}, \emph{randomness}, and \emph{complexity} form  a conceptual \emph{triangular} landscape.  Table~\ref{tab:axis_meaning} summarizes the interpretive role of each axis.

\begin{table*}
\footnotesize
\caption{\label{tab:axis_meaning} Interpretive meaning of the three conceptual axes used throughout this paper.}
\begin{ruledtabular}
\begin{tabular}{ll}
\textbf{Axis} & \textbf{Meaning} \\
\hline
Regularity & Predictable, compressible structure (periodicity, symmetries, rules) \\
Randomness & Pure unpredictability (noise, stochasticity) \\
Complexity & Non-trivial structure that is neither fully regular nor fully random \\
\end{tabular}
\end{ruledtabular}
\end{table*}

Real–world datasets seldom sit at the extremes; instead they occupy an interior region of this conceptual landscape, blending structure and noise to varying degrees. Different measures provide insights into distinct aspects of this landscape—some quantify uncertainty, others characterize structural regularity, and some assess complexity by distinguishing meaningful patterns from noise.
 The placement of measures in Table~\ref{tab:landscape} therefore serves as a conceptual guide rather than a precise coordinate system.

\begin{table*}
    \footnotesize
    \caption{\label{tab:landscape} \textbf{Conceptual Landscape — Dominant Emphasis of Each Measure}. An entry of “–” means the measure does not directly quantify that axis.}
    \begin{ruledtabular}
        \begin{tabular}{lccc}
            \textbf{Measure} & \textbf{Regularity} & \textbf{Randomness} & \textbf{Complexity} \\
            \hline
            \multicolumn{4}{c}{\textit{Static (Entropy‑based) Measures}} \\ \hline
            Shannon Entropy~\cite{Shannon1948} & – & \checkmark & – \\
            Rényi / Tsallis Entropies~\cite{Renyi1961,Tsallis2003} & – & \checkmark & – \\
            Approx. Entropy (ApEn)~\cite{Pincus1991} & weak & \checkmark & weak \\
            Sample Entropy (SampEn)~\cite{Richman2000} & weak & \checkmark & weak \\
            Permutation Entropy~\cite{Bandt2002} & weak & \checkmark & weak \\ \hline
            \multicolumn{4}{c}{\textit{Algorithmic / Structural Measures}} \\ \hline
            Kolmogorov Complexity~\cite{Kolmogorov1968,Chaitin1969} & \checkmark & \checkmark & \checkmark \\
            Logical Depth~\cite{Bennett1988} & \checkmark & – & \checkmark \\
            Effective Complexity~\cite{Gellman2004} & \checkmark & – & \checkmark \\
            Statistical Complexity~\cite{Crutchfield2017} & \checkmark & – & \checkmark \\
            Sophistication~\cite{Koppel1987} & \checkmark & – & \checkmark \\ \hline
            \multicolumn{4}{c}{\textit{Dynamical Measures}} \\ \hline
            Entropy Rate~\cite{Porta2001} & \checkmark & \checkmark & approx. \\
            Transfer Entropy~\cite{Schreiber2000} & \checkmark & \checkmark & some \\
            Active Information Storage~\cite{Wibral2013} & \checkmark & \checkmark & some \\
            Information Modification~\cite{Lizier2010} & \checkmark & – & \checkmark \\
        \end{tabular}
    \end{ruledtabular}
\end{table*}

\textit{Reading the table:} Rows with a check under \emph{Randomness} (e.g., Shannon entropy) measure uncertainty but ignore deeper structure.  
Algorithmic measures (middle block) capture rich organization and thus appear under both \emph{Regularity} and \emph{Complexity}.  
Dynamical measures straddle axes because they probe both predictability (regularity) and novelty (randomness) in time‑evolving signals. Some measures, such as Kolmogorov Complexity, capture all three properties simultaneously, making them broadly applicable but challenging to compute precisely. Understanding each measure’s emphasis clarifies why two metrics may deliver different ``complexity'' verdicts on the same dataset and helps analysts choose the appropriate tool.

\subsection{Computational Depth vs.\ Practical Accessibility}
Conceptual richness is only half the story. In practice we must ask whether a measure is \textit{feasible} to estimate. Greater conceptual depth generally comes with poorer accessibility—a trade‑off that motivates approximate, data‑driven surrogates. Table~\ref{tab:depth_access_table} arranges the same measures on two axes of \textbf{Computational Depth} (theoretical richness) and \textbf{Practical Accessibility} (ease of estimation).

\begin{table*}
    \footnotesize
    \caption{\label{tab:depth_access_table} \textbf{Computational Depth vs. Practical Accessibility}. ``Easy'' means estimable from modest data with standard software; ``Hard'' means requiring large data, specialized estimators, or is formally uncomputable.}
    \begin{ruledtabular}
        \begin{tabular}{lccc}
            \textbf{Measure} & \textbf{Depth} & \textbf{Accessibility} & \textbf{Computability / Approximation} \\
            \hline
            Shannon Entropy~\cite{Shannon1948}                       & Shallow & Easy & Yes (directly via histogram) \\
            Rényi/Tsallis Entropies~\cite{Renyi1961,Tsallis2003}     & Shallow & Easy & Yes (for discrete data; sensitive to $\alpha/q$ in continuous cases) \\
            ApEn, SampEn~\cite{Pincus1991,Richman2000}               & Moderate & Easy & Yes (parametric estimator, sensitive to data size) \\
            Permutation Entropy~\cite{Bandt2002}                     & Moderate & Easy & Yes (ordinal pattern statistics) \\
            Transfer Entropy~\cite{Schreiber2000}                    & Moderate & Moderate (data-hungry) & Approximable (embedding $\&$ estimator-dependent) \\
            Entropy Rate~\cite{Porta2001}                            & Moderate & Moderate (estimator bias) & Approximable via Markov models, NN estimators \\
            Active Information Storage~\cite{Wibral2013}             & Moderate & Moderate & Approximable via time-series embedding + MI estimators \\
            Statistical Complexity~\cite{Crutchfield2017}            & Deep & Moderate (requires $\epsilon$-machine reconstruction) & Approximable via model-driven inference \\
            Effective Complexity~\cite{Gellman2004}                  & Deep & Moderate (model choice) & No (approximable via MDL, Bayesian methods) \\
            Information Modification~\cite{Lizier2010}               & Deep & Hard (immature estimators) & Approximable (PID-based methods; evolving estimators) \\
            Logical Depth~\cite{Bennett1988}                         & Deep & Inaccessible & No (no reliable approximation framework) \\
            Sophistication~\cite{Koppel1987}                         & Deep & Inaccessible & No (theoretical concept, lacks stable approximation method) \\
            Kolmogorov Complexity~\cite{Kolmogorov1968,Chaitin1969}  & Deepest & Inaccessible (uncomputable) & No (approximated via compression algorithms) \\
        \end{tabular}
    \end{ruledtabular}
\end{table*}

\subsection{Why Computability Matters—and How We Approximate}
In theory, many fundamental complexity measures — notably \textbf{Kolmogorov complexity}~\cite{Kolmogorov1968,Chaitin1969}, \textbf{Effective complexity}~\cite{Gellman1996}, and \textbf{Logical Depth}~\cite{Bennett1988} — provide deep insights into regularity, randomness, and complexity. However, in practice:

\begin{itemize}
    \item Some measures are \textbf{exactly computable} from finite data (e.g., Shannon Entropy),
    \item Others are \textbf{approximable} with bias‑controlled estimators (e.g., Entropy Rate, Transfer Entropy, AIS)~\cite{Kennel2005,Wibral2013},
    \item The most structural—Kolmogorov, Effective Complexity, Logical Depth, Sophistication—are \textbf{formally uncomputable} or infeasible; we approximate them via compression ratios, MDL bounds, or learned latent models,
    \item Some are \textbf{currently intractable or poorly estimated} (e.g., Information Modification).
\end{itemize}


Modern machine learning approaches learn approximations to: compressibility, structure extraction, and noise disentanglement, implicitly approximating the intractable measures without computing them formally.
Modern data‑driven methods serve as \textbf{computational prostheses} for these otherwise inaccessible or partially accessible quantities:

\begin{itemize}
    \item \textbf{Autoencoders/VAEs}~\cite{Alemi2019,Kingma2014} learn minimal latent codes, approximating compressibility and entropy.
    \item \textbf{Symbolic regression with sparsity regularisers}~\cite{Brunton2016,Cranmer2020} searches for short, interpretable generative rules, echoing Effective Complexity.
    \item \textbf{Physics‑informed neural networks (PINNs)}~\cite{Raissi2019,Karniadakis2021} discover governing equations, acting as operational proxies for deep structural measures.
\end{itemize}

By casting inference as large-scale optimization of description length, deep learning cannot sidestep formal uncomputability, but it provides practical surrogates, allowing partial estimation of key patterns that reside in the core of complexity properties ~\cite{Hinton2006,Goodfellow2014}. These approximation techniques come with inherent limitations and biases, which arise from estimator choice, dataset size, sampling strategy, and assumptions embedded in learned models.

\paragraph*{Key Take‑Home}
For foundational measures the pivotal question is no longer \emph{“Is it computable?”} but \emph{“How well can we approximate it—and with what bias?”}. Knowing where a measure sits—both in the \emph{Regularity–Randomness–Complexity} landscape and on the \emph{Depth–Accessibility} plane—guides practitioners in selecting tools suited to their data, scientific goals, and computational budget.

\section{Discussion}
\subsection{Motivations and Conceptual Synthesis}
Complexity science has accumulated a rich catalog of measures—statistical, algorithmic, and dynamical—yet their operational scopes have remained fragmented for characterization of complexity in empirical data.~\cite{Lloyd2001,Prokopenko2009,Ladyman2013}.
By recasting those measures within three orthogonal, but inter‑dependent, axes

By organizing these measures along three axes of a conceptual landscape, we create a framework that clarifies their operational meaning and limitations:
\begin{itemize}
    \item \textbf{Regularity} (structured predictability and compressibility),
    \item \textbf{Randomness} (unpredictability and irreducibility), and
    \item \textbf{Complexity} (non-trivial interplay between structure and disorder),
\end{itemize}

The taxonomy of complexity measures reveals several key insights:
\begin{itemize}
    \item \textbf{Statistical entropy measures} (e.g., Shannon, Renyi, Tsallis) primarily focus on randomness through calculation of unpredictability; they ignore deeper regularities unless augmented by multiscale extensions that capitalize on coarse-graining.
    \item \textbf{Algorithmic complexity measures} (e.g., Kolmogorov complexity, Effective Complexity, Logical Depth) simultaneously attempt to quantify regularity and randomness in order to define complexity at their boundary, but their power is offset by formal uncomputability.
    \item \textbf{Dynamical information measures} (e.g., Entropy Rate, Transfer Entropy, Active Information Storage) explicitly aim to probe not just \textit{what} information is present in data but \textit{how} it transforms in time.
\end{itemize}

\paragraph*{Computability as the pivot.}
Our depth–accessibility analysis shows a systematic trade‑off: the deeper a measure specifies structure, the less accessible it becomes. Some metrics (Shannon, permutation entropy) are point‑and‑compute; others (Kolmogorov, Logical Depth) remain theoretical ideals, yet can be partially approximated through compression ratios or learned latent representations

\textbf{Conceptual maps as a toolkit.}
\begin{enumerate}
    \item \textbf{Regularity–Randomness–Complexity triangle} organizes measures based on their emphasis.
    \item \textbf{Depth-×-Accessibility plane} highlights estimation difficulty.
    \item \textbf{Computability flow chart} classifies exact, approximate, and uncomputable regimes.
\end{enumerate}
Together these maps let researchers choose a metric that fits both their scientific question and their computational budget. Having established this foundation, we now turn to the relationship between these classical complexity notions and the emerging field of data-driven modeling using machine learning and latent space discovery.

\subsection{Bridging Classical Measures and Modern Data-Driven Methods}

Classical measures of regularity, randomness, and complexity offer profound conceptual insights, yet many are either formally uncomputable or practically fragile~\cite{Li2019,Bennett1988,Crutchfield2012}.  
Kolmogorov complexity is uncomputable, Effective Complexity depends on model class specification, Logical Depth is challenging to approximate, and even measures like Transfer Entropy require data-hungry, bias-controlled estimators.

This tension — between the depth of theoretical constructs and the limits of computation — is not merely a historical curiosity. It is precisely here that \textbf{modern data-driven methods have found their role}.
Increasing availability of high-dimensional, complex data has motivated the development of \textbf{machine learning techniques} that—often implicitly—approximate exactly the quantities that classical theory defines but cannot compute directly. These modern approaches leverage compression algorithms, variational inference, and latent-space embeddings to extract meaningful structure from data without needing explicit formulations of complexity.

Rather than formally computing uncomputable quantities, modern machine learning approximates their operational goals, serving as pragmatic surrogates for classical complexity measures (Table \ref{tab:classical_modern_bridge}):
\begin{itemize}
    \item \textbf{Compression:} \textbf{\textit{Autoencoders}} and \textbf{\textit{Variational Autoencoders (VAEs)}}, learn latent representations that preserve meaningful structure while discarding noise~\cite{Hinton2006,Kingma2014}. While they do not compute Kolmogorov Complexity explicitly, their learned representations align with principles of efficient encoding in minimum description length (MDL) theory, preserving meaningful structure while discarding noise.
    \item \textbf{Regularity Extraction:} \textbf{Physics-Informed Neural Networks (PINNs)} enforce known physical laws within learning~\cite{Raissi2019,Karniadakis2021}, constraining models toward structured behavior. This aligns with the goals of \textbf{\textit{Effective Complexity}}, which quantifies rule-based organization over randomness.
    \item \textbf{Entropy and Density Modeling:} Variational models such as VAEs and \textbf{normalizing flows} estimate probability densities in latent spaces~\cite{Rezende2015,Kingma2014}, offering approximations of \textbf{\textit{statistical complexity}} and predictive uncertainty.
    \item \textbf{Learning Dynamical Regularities:} \textbf{Latent ODEs}~\cite{Chen2018,Rubanova2019}, \textbf{Koopman autoencoders}~\cite{Lusch2018,Otto2019}, and related approaches seek to find \textbf{low-dimensional and/or locally linear dynamical systems} that explain high-dimensional observed trajectories. These methods operationalize the compression of complex temporal dependencies into structured, learnable flows — analogous to capturing regularities in \textbf{entropy rates} and \textbf{information storage}.     
    \item \textbf{Symbolic Regression:} Symbolic regression methods uncover minimal algebraic descriptions of system dynamics~\cite{Schmidt2009,Udrescu2020,Champion2019}. While not a direct measure of Logical Depth, they capture structured dependencies where simple rules unfold into complex behavior
\end{itemize}

Modern data-driven methods function as applied approximations to classical complexity ideals. While framed in terms of optimization and learning rather than formal complexity theory, they navigate the same landscape: \textbf{capturing regularity}, \textbf{discarding randomness}, and \textbf{representing complexity efficiently}.

These models do not compute quantities like Kolmogorov complexity or Effective Complexity in a strict sense, but they approximate their \textit{operational goals}: compressing representations, discovering governing rules, and distinguishing meaningful structure from noise—guided by objectives like reconstruction loss, predictive likelihood, or sparsity constraints.

Latent representations in modern models are not just engineering conveniences; they encode the very trade-offs that complexity theory formalizes, adapting to real-world constraints. Their empirical success suggests that real-world data manifolds possess sufficient structure to make \textbf{\textit{approximate forms of ``Complexity'' learnable}}.

\begin{table*}
\footnotesize
\caption{\label{tab:classical_modern_bridge}\textbf{Classical Concepts, Deep Features, and Modern Approximations}. Each modern method approximates the operational goal that classical theory defines but cannot compute directly.}
\begin{ruledtabular}
\begin{tabular}{lcc}
\textbf{Classical Concept} & \textbf{Operational Goal (Deep Feature)} & \textbf{Modern Approximation} \\
\hline
\multicolumn{3}{c}{\textit{Compression and Structure}} \\ \hline
Kolmogorov Complexity & Minimize description length, extract structure & Autoencoders, VAEs; Lempel–Ziv proxies \\
Effective Complexity  & Identify regularities through efficient encoding & Structure-aware latent models (e.g., $\beta$-VAEs) \\
\hline
\multicolumn{3}{c}{\textit{Rule Discovery and Regularity Extraction}} \\ \hline
Logical Depth         & Discover governing rules, symmetries & Sparse symbolic regression (SINDy, Eureqa) \\
Symmetry Discovery & Identify invariant properties & Physics-informed neural networks (PINNs) \\
\hline
\multicolumn{3}{c}{\textit{Predictive and Dynamical Structure}} \\ \hline
Entropy Rate          & Optimize sequence predictability; minimize redundancy & Latent ODEs; Koopman autoencoders; RNN forecasters \\
Active Information Storage & Preserve information over time & Recurrent autoencoders; gated RNNs \\
\hline
\multicolumn{3}{c}{\textit{Handling Randomness and Uncertainty}} \\ \hline
Statistical Entropy   & Discard noise, focus on signal & Normalizing flows; variational inference \\
\hline
\multicolumn{3}{c}{\textit{Emergent or Higher-Order Complexity}} \\ \hline
Information Modification & Discover non-trivial synergies & Attention mechanisms?\footnote{Though not explicit, we speculate about this possible interpretation. Recent studies analyze attention weights in transformers as multi‑source interaction patterns that go beyond additive information, making them empirical probes of synergistic (information‑modification) effects~\cite{Rosas2020,Lizier2010}.  Including these architectures therefore illustrates how higher‑order complexity—once purely theoretical—can now be approximated in large‑scale models.} \\
\end{tabular}
\end{ruledtabular}
\end{table*}

\subsection{Toward Operational Complexity in Latent Spaces}
Latent-space models represent a methodological shift—rather than \emph{computing} complexity measures explicitly, they \emph{learn} structured mappings from high-dimensional observations to compact latent codes in which \textbf{regularity}, \textbf{randomness}, and \textbf{complexity} are implicitly separated. These codes form an \emph{operational domain}, allowing structured information to be efficiently isolated, compressed, and, in some cases, represented via symbolic formulations.

Several key trends illustrate this convergence:

\subsubsection{\textbf{Compression and Structure Extraction via Autoencoders}}
Autoencoders perform explicit compression by mapping observations into a lower‑dimensional latent code and reconstructing them, capturing essential regularities while discarding noise~\cite{Hinton2006,Vincent2008}.

\textbf{Autoencoders} perform explicit compression:

\begin{itemize}
    \item \emph{Encoder}: $x \mapsto z$ maps an input $x$ to a lower‑dimensional latent vector $z$.
    \item \emph{Decoder}: $z \mapsto \hat{x}$ reconstructs the original data $\hat{x}$ from the latent vector $z$.
\end{itemize}

With appropriate regularization—sparsity penalties, noise injection, or a narrow bottleneck—$z$ retains the essential regularities needed for accurate reconstruction while discarding unimportant variability~\cite{Vincent2008}.  In classical terms, $z$ approximates a \emph{compressed description} of the data.

\textbf{Variational Autoencoders (VAEs)} extend this idea by enforcing \emph{a probabilistic latent prior} and adding a Kullback–Leibler term that trades off reconstruction accuracy against latent entropy~\cite{Kingma2014,Alemi2019}.  The regularization parameter $\beta$ in $\beta$‑VAEs explicitly controls this trade‑off \cite{Higgins2017}, allowing a continuum between faithful reconstruction and stronger compression.

From the perspective of classical complexity theory:
\begin{itemize}
    \item The latent vector $z$ is an empirical proxy for a minimal description length, echoing \textbf{Kolmogorov} and \textbf{Effective Complexity}.
    \item The decoder implements the \emph{unfolding computation} from code to data, paralleling \textbf{Logical Depth}: longer or more nonlinear decoders reflect greater computational effort to recreate structure.
\end{itemize}

\subsubsection{\textbf{Dynamics in Latent Spaces: Toward Predictive Regularity}}
Modern latent compression extends beyond static data to modeling \textbf{temporal sequences}, enabling compact representations of evolving structures. 

\begin{itemize}
    \item \textbf{Latent Ordinary Differential Equations (Latent ODEs)}~\cite{Chen2018,Rubanova2019} learn continuous‑time dynamics in a compressed latent space, capturing predictive regularities by structuring latent dynamics to reduce uncertainty along trajectories, effectively minimizing entropy rates.
    \item \textbf{Koopman Autoencoders}~\cite{Lusch2018,Otto2019} learn latent coordinates in which the dynamics are approximately \emph{linear}, leveraging Koopman operator theory~\cite{Koopman1931,Koopman1932} to model nonlinear observations.
\end{itemize}

Both models operationalize dynamical complexity: they encode evolution in a low‑entropy latent manifold, where the latent dynamical models serve as \textbf{compressed, structured approximations} paralleling the goals of \textbf{Effective Complexity} and \textbf{Entropy Rate}.

\subsubsection{\textbf{Symbolic Latent Representations: Unfolding Logical Depth}}
Symbolic‐regression methods such as \textbf{Eureqa}~\cite{Schmidt2009} and \textbf{SINDy}~\cite{Brunton2016,Champion2019} extract explicit algebraic rules for latent dynamics. In this context:

\begin{itemize}
    \item Latent variables become \textbf{discoverable symbolic coordinates}; recovered equations provide interpretable structure.
    \item The process of expanding concise symbolic rules into detailed trajectories reflects \textbf{Logical Depth}, where computational effort is required to transform minimal descriptions into full emergent behavior.
\end{itemize}

Symbolic latent models therefore approach the ideal of a \emph{minimal sufficient generative model}: a compact symbolic program that reproduces the observed phenomena.

\subsubsection{\textbf{Physics-Informed Latent Learning: Anchoring Regularity}}
\textbf{Physics‑Informed Neural Networks (PINNs)} and \textbf{Neural Operators} such as DeepONet embed governing equations or conservation laws into the training loss, steering the latent space toward physically consistent manifolds~\cite{Raissi2019,Karniadakis2021,Lu2021,Wang2021}.  

By embedding \emph{structural priors} into training, these models guide latent representations toward \emph{meaningful regularities} rather than mere observed pattern replication. This approach mitigates a key limitation of unconstrained compression, in which the model may optimize reconstruction error yet ignore the underlying physical structure.

\subsubsection{Limitations of Neural Proxies}
While Autoencoders, VAEs, latent ODEs, and PINNs offer operational surrogates for uncomputable complexity measures, they each introduce specific inductive biases and failure modes that affect their reliability.

\paragraph{VAEs}  
Variational autoencoders optimize the evidence‑lower‑bound (ELBO) objective rather than an explicit description‑length criterion that underlies Kolmogorov Complexity~\cite{Alemi2019}. The common use of an isotropic Gaussian prior in VAEs inadequately captures rare or multimodal patterns. This results in oversmoothed reconstructions, leading to an underestimation of structured complexity, particularly in heavy-tailed data distributions~\cite{Casale2018}.

\paragraph{Latent ODEs under Noise and Misspecification}  
Latent ODE models enforce smooth continuous‑time dynamics learned from data~\cite{Chen2018}.  
In the presence of observational noise or abrupt regime shifts, the learned vector field may oversmooth true dynamics, biasing the latent representation. Noise can also push latent states into regions where numerical integration error accumulates, causing long‑term divergence from ground‑truth trajectories~\cite{Rubanova2019}.

\paragraph{PINNs and Prior Misalignment}  
Physics‑informed neural networks combine a data‑fit loss with a partial‑differential‑equation residual~\cite{Raissi2019}.  The balance between these losses is set by hand‑tuned hyperparameters. If the assumed PDE omits relevant physics, PINNs can either overfit to noisy observations or enforce incorrect constraints, leading to physically inconsistent reconstructions and misestimated dynamical complexity.~\cite{Karniadakis2021}.

\subsubsection{Decision Guide for Proxy Selection}

\paragraph{Measure Pluralism and Domain Alignment.}
Complexity takes different forms across domains—from thermodynamic assembly in biophysics to algorithmic depth in computation—making it impossible for a single measure to capture all aspects. Practitioners must therefore choose a proxy whose foundational assumptions (coarse‑graining, resource model, dynamical regime) match the physics and data constraints of the target system. The following decision guide outlines key considerations (Table \ref{tab:proxy_selection}):

 \begin{table*}
\footnotesize
\caption{\label{tab:proxy_selection} \textbf{Decision Guide for Proxy Selection}: Key considerations and recommended approximation strategies.}
\begin{ruledtabular}
\begin{tabular*}{\textwidth}{@{\extracolsep{\fill}} l p{4cm} p{8cm} }
\textbf{Step} & \textbf{Key Consideration} & \textbf{Recommended Approach} \\
\hline
\textbf{Step 1} & Nature of Structure & If strong \textit{a priori} knowledge exists (e.g., PDEs, conservation laws), use \textbf{Physics-Informed Neural Networks (PINNs)} or \textbf{symbolic latent models}. Otherwise, proceed to Step 2. \\
\textbf{Step 2} & Data Dimensionality and Dynamics & 
\begin{minipage}[t]{8cm}
    \begin{itemize}
        \item For \textbf{time-series with continuous dynamics} and moderate noise: Use \textbf{Latent ODEs} with noise-robust encoders and validate on synthetic trajectories.
        \item For \textbf{high-dimensional static data or spatial images}: Use \textbf{Variational Autoencoders (VAEs)} with expressive priors (e.g., VampPrior, mixture models). Otherwise, proceed to Step 3.
    \end{itemize}
\end{minipage} \\
\textbf{Step 3} & Interpretability vs. Compression Fidelity & 
\begin{minipage}[t]{8cm}
    \begin{itemize}
        \item If \textbf{interpretability} is critical (especially in small samples): Use \textbf{Compression-Based Proxies} (e.g., LZMA, BZIP2) to approximate Kolmogorov complexity.
        \item If \textbf{flexible latent representations} are preferred over strict interpretability: Use \textbf{Autoencoders or InfoMax Models} (e.g., InfoVAE) for richer embedding spaces.
    \end{itemize}
\end{minipage} \\
\textbf{Step 4} & Computational Budget and Scalability & 
\begin{minipage}[t]{8cm}
    \begin{itemize}
        \item For \textbf{large datasets} where deep latent models are feasible: Use \textbf{VAEs or Latent ODEs} with mini-batch training and early stopping.
        \item For \textbf{resource-constrained environments} or rapid prototyping: Use \textbf{Compression-Based Proxies} or \textbf{AIS/TE Estimators} (for dynamical data) with efficient \(k\)-NN or histogram methods.
    \end{itemize}
\end{minipage} \\
\textbf{Step 5} & Validation and Benchmarking & Regardless of choice, always \textbf{benchmark your proxy} on synthetic ground-truth systems (e.g., logistic map, Lorenz attractor, symbolic sequences) to quantify bias and variance. Supplement metrics with qualitative inspection of reconstructions or decompressed outputs. \\
\end{tabular*}
\end{ruledtabular}
\end{table*}

\noindent\emph{Practical recommendation.} In addition to the steps above, consider richer priors (hierarchical VAEs) or hybrid discrete–continuous latent models when the chosen proxy underestimates structured variability. Systematic benchmarking on synthetic models remains the most reliable strategy for detecting and correcting proxy biases, ensuring closer alignment between latent complexity and theoretical principles.

\subsubsection{\textbf{The Latent Space as an Operational Arena for Complexity Negotiation}}
Latent spaces are more than mathematical conveniences—they serve as \emph{operational arenas} where learning systems structure data through three interconnected axes of complexity.

\begin{itemize}
    \item \textbf{Regularity extraction}: How much structure can be compressed?
    \item \textbf{Randomness modeling}: How much stochasticity must be retained or explicitly modeled?
    \item \textbf{Mapping complexity}: How intricate is the transformation between latent codes and observations (logical depth, ``predictive'' entropy)?
\end{itemize}

By learning such spaces, neural networks implement—approximately and at scale—the decomposition of regularity, randomness, and complexity~\cite{Tishby2000,Tishby2015,Achille2018} that classical theory could only formalize.  They learn compressed, structured, and predictive representations from high‑dimensional noisy data without assuming a fully specified model \emph{a priori}. Thus, operational complexity in latent spaces bridges long-standing theoretical ambitions with new frontiers in scalable learning.

\section{Conclusion}
True complexity cannot be captured through isolated calculations of entropy or predictability alone—it requires a principled framework for understanding the fundamental properties of data and their interactions. We framed the landscape in terms of three axes—\textbf{regularity}, \textbf{randomness}, and \textbf{complexity}—and classified statistical, algorithmic, and dynamical measures accordingly.  

Three conceptual maps summarize that classification:

\begin{itemize}
    \item the \emph{Regularity–Randomness–Complexity} triangle;
    \item the \emph{Depth–Accessibility} plane, which exposes the trade‑off between conceptual richness and computational feasibility;
    \item a \emph{Computability flow} separating exact, approximate, and uncomputable regimes.
\end{itemize}
Our taxonomy reveals both the distinct operational meanings of complexity measures and the deeper challenge of computability—where fundamental structures like Kolmogorov complexity and Effective Complexity remain uncomputable, requiring pragmatic approximations. Against this backdrop, we highlighted how \textbf{modern machine learning methods} —including autoencoders, variational models, latent ODEs, symbolic regression, and physics-informed networks —can be understood as \textbf{operational approximations} to classical complexity-theoretic ideals. Latent spaces emerge as operational arenas where compressibility, regularity extraction, and complexity negotiation converge, enacting classical theoretical ambitions in practice. Thus, latent spaces are more than just lower-dimensional embeddings—they are operational domains where complexity is actively discovered, structured, and negotiated in modern learning systems.

\paragraph*{Outlook.}  
Future learning systems should explicitly integrate compressibility, regularity extraction, and complexity management as core objectives—both in theoretical models and data-driven approaches. Key directions include:

\begin{itemize}
    \item defining quantitative metrics of \emph{latent‑space complexity};
    \item embedding compression‑theoretic objectives directly in learning algorithms;
    \item linking formal notions of regularity and randomness to training dynamics.
\end{itemize}

Such advances would tighten the connection between complexity theory and data‑driven discovery, enabling systems that not only fit data but \emph{understand} it structurally, compress it meaningfully, and predict it reliably.  
The resulting synthesis has implications for complexity science, machine learning, and the emerging interfaces of \emph{physics for AI} and \emph{AI for physical systems}, where physically informed architectures can leverage precise operational definitions of regularity, randomness, and complexity to improve efficiency and interpretability.

On one hand, insights from complexity, regularity, and compression can inform the design of \textbf{physics-inspired architectures} — neural networks that respect conservation laws, symmetries, and structural constraints to achieve greater efficiency and interpretability. On the other hand, advances in \textbf{data-driven modeling of physical systems} — from fluid turbulence to biological dynamics — stand to benefit from a clearer operational understanding of how regularities are extracted, randomness is managed, and complexity is negotiated within latent spaces.

By bridging classical theory and modern methods, we hope this work sparks dialogue and lays the foundation for future advances at the intersection of complexity science, physics, and artificial intelligence.


\nocite{*}
\bibliographystyle{plain}
\bibliography{CxRx}

\end{document}